\documentclass[runningheads]{llncs}
\usepackage[T1]{fontenc}
\usepackage{graphicx} % Required for inserting images
\usepackage[table]{xcolor}
\usepackage{listings}
\usepackage{hyperref}
\usepackage{enumitem}
\usepackage{booktabs}
\usepackage{multicol}
\usepackage{xurl}
\usepackage[htt]{hyphenat}
\usepackage{breakcites}
\usepackage{multirow}

\definecolor{codegreen}{rgb}{0,0.6,0}
\definecolor{codegray}{rgb}{0.5,0.5,0.5}
\definecolor{codepurple}{rgb}{0.58,0,0.82}
\definecolor{backcolour}{rgb}{0.95,0.95,0.92}

\lstdefinelanguage{json}{
    basicstyle=\normalfont\ttfamily,
    numbers=left,
    numberstyle=\scriptsize,
    stepnumber=1,
    numbersep=8pt,
    showstringspaces=false,
    breaklines=true,
    frame=lines,
    backgroundcolor=\color{backcolour},
    literate=
     *{0}{{{\color{black}0}}}{1}
      {1}{{{\color{black}1}}}{1}
      {2}{{{\color{black}2}}}{1}
      {3}{{{\color{black}3}}}{1}
      {4}{{{\color{black}4}}}{1}
      {5}{{{\color{black}5}}}{1}
      {6}{{{\color{black}6}}}{1}
      {7}{{{\color{black}7}}}{1}
      {8}{{{\color{black}8}}}{1}
      {9}{{{\color{black}9}}}{1}
      {:}{{{\color{codegreen}{:}}}}{1}
      {,}{{{\color{codegreen}{,}}}}{1}
      {\{}{{{\color{black}{\{}}}}{1}
      {\}}{{{\color{black}{\}}}}}{1}
      {[}{{{\color{black}{[}}}}{1}
      {]}{{{\color{black}{]}}}}{1},
}

\lstdefinestyle{mystyle}{
    backgroundcolor=\color{backcolour},
    commentstyle=\color{codegreen},
    keywordstyle=\color{magenta},
    numberstyle=\tiny\color{codegray},
    stringstyle=\color{codepurple},
    basicstyle=\ttfamily\footnotesize,
    breakatwhitespace=false,
    breaklines=true,
    captionpos=b,
    keepspaces=true,
    numbers=left,
    numbersep=5pt,
    showspaces=false,
    showstringspaces=false,
    showtabs=false,
    tabsize=2
}

\begin{document}

\title{Towards a foundational model for recognising diastematic Gregorian notation}
%\title{A unified model for recognising \\ Gregorian staff notations}
\titlerunning{Foundational Grego}
\authorrunning{Kurek and Hajič}
%\author{Pau*, Jirka*, Carles, Martina, Marketa, Gerard, Vojta, Samuel, Jan, Alicia}

\author{Daniel Kurek\inst{1} \and Jan~Haji\v{c} jr.\inst{1}}
%\authorrunning{P. Torras et al.}
\institute{Institute of Formal and Applied Linguistics, Charles University, Prague, Czechia  \\ \email{hajicj@ufal.mff.cuni.cz}
%\and Moravian Library, Brno, Czechia%, \email{xxx} 
}

\date{May 2026}

\maketitle

\begin{abstract}
    Optical recognition of Gregorian notation has recently been attempted with end-to-end methods, with four datasets introduced. However, each of these datasets is in a different encoding. We design a common encoding based on the S-GABC proposal, convert all four datasets to this common encoding, and train a shared end-to-end foundational model for diastematic Gregorian notation that establishes a new state of the art across all four datasets.
\end{abstract}

\section{Introduction}

Gregorian chant, the liturgical monody of the Latin church, was the first musical tradition to be notated on a staff, already in the 11th century \cite{atkinson2008critical}. It is also one of the largest homogeneous musical traditions, with up to 30.000 extant manuscripts \cite{helsen2014chantOmr}, and it has been a major pillar of European musical identity and an essential element of its soundscape up until the late 20th century \cite{hiley1993western}. Despite significant loss of status since the 2nd Vatican Council and the subsequent liturgical reforms in the 1970s, it is still practised not only in Europe, but in all places with Catholic communities worldwide. Within musicology, chant scholarship is a large field, with its own multi-day conference.\footnote{\url{https://cantusplanus.musicology.org/evora-2026/}} 

But while the scale of chant presents major opportunities for computational musicology to have an impact \cite{cornelissen2020studying,cornelissen2020mode,helsen2021sticky,hajic2023,lanz2023,eipert2026communities,HajicJr2025genomeOfMelody,lanz2025gregorianMelody,hajic2025unseen,dvorakova2026makingChantComputingEasy}, this is held back by the need for correspondingly large-scale datasets. Specifically for melody research, transcriptions are necessary (typically in the Volpiano \cite{helsen2011volpiano,lacoste2012cantus}, GABC \cite{cornelissen2020studying} or MEI \cite{Helsen2017morphologyMEI,DeLuca2019capturingNeumesMEI} encodings), and these are time-consuming and require highly specialised expertise in reading chant notations. While over 40,000 fully transcribed melodies are present in the Cantus network of databases \cite{cornelissen2020studying,dvorakova2026makingChantComputingEasy}, these in fact mostly come from just several fully transcribed manuscripts: full-size Gregorian chant books contain 1000-4000 melodies each.

It is thus no surprise that optical music recognition (OMR) for Gregorian chant has seen multiple systematic efforts to tackle it \cite{helsen2014chantOmr,vigliensoni2019image,hartelt2024optical,wick2019ommr4all,fujinaga2019single,fuentes-martinezAlignedMusicNotation2026,martinez-sevillaTowardsUniversalOptical2024,Bouressa2025pixelsToPaleography}, with some success. However, the available datasets are split across many different encodings --- sometimes obviously distinct, sometimes subtly so. Despite the principles of Gregorian staff notation being remarkably consistent across its multiple visual realisations (square notes vs. rhombic notes, subtle graphics vs. thick lines, etc.), it is thus not straightforward to pool these resources.

\section{Related work}

While the notation of Gregorian chant may seem peripheral to the justification of OMR in terms of user-facing applications, it has in fact seen a lot of OMR research. Besides didactic reasons, the justification comes from the needs of digital and computational humanities: with tens of thousands of extant manuscripts, containing some 1000-3000 chants each, computational methods are the only viable way of robustly understanding this major chapter of European cultural heritage --- and these in turn require machine-readable data, where again the scale of chant precludes any path but automation, via OMR. At the same time, the staff notation of chant is relatively straightforward: at least from the 13th century onwards, chant melodies can be conceptualised simply as sequences of pitches, and their alignment to text (usually expressed in terms of syllable and word boundaries). 

%This conceptual simplicity made the notation assembly stage of the traditional OMR pipeline \cite{rebeloOMRStateOfTheArt2012,calvo-zaragozaUnderstandingOpticalMusic2021} trivial for chant, making it attractive also from the engineering point of view.

Chant was first substantially addressed with the interactive Gamera toolkit \cite{droettboom2003gamera}, based on kNN classification of neume shapes after staff removal. This gradually evolved into web-based tools Pixel.js \cite{saleh2017pixel}, Neon \cite{burlet2012neon} and Neon2 \cite{regimbal2019neon2}, and the Rodan framework for orchestrating jobs with interactive steps \cite{hankinson2014rodan}.

A different software environment, a monolithic application instead of the Rodan-orchestrated combined workflows, was developed by the Corpus Monodicum project: the OMMR4All software \cite{wick2019ommr4all,hartelt2024optical}. However, it follows a similar workflow (staff detection, symbol detection, notation assembly, and final encoding for the MonodiKit environment \cite{eipert2023monodikit}), though the text recognition component of the pipeline does use an image-to-sequence model \cite{hartelt2024optical}.

Most recently, end-to-end recognition with a host of different image-to-sequence models was introduced as the aligned music notation and lyrics transcription (AMNLT) task by Fuentes-Martinez et al. \cite{fuentes-martinezAlignedMusicNotation2026}. This is the first substantial work on OMR for chant using end-to-end methods such as the Sheet Music Transformer \cite{rios-vilaEndtoendOpticalMusic2023,RiosVila2024sheetMusicTransformer}. It introduces four distinct datasets for this purpose (GregoSynth, Einsiedeln, Salzinnes, Solesmes) with ground truth suitable for staff-wise recognition, and measures results on these datasets for 11 different models. We follow up on this work: we take the same datasets and evaluation procedure, harmonise the encodings to enable merging the datasets into one, and train shared models that substantially improve on this work, opening the doors towards a foundational model for diastematic Gregorian notation.

\section{Datasets}

The baseline work \cite{fuentes-martinezAlignedMusicNotation2026} has combined four datasets of diastematic (that is, staff) Gregorian notation. The dataset sizes are summarized in table \ref{tab:datasets}. The datasets are available via HuggingFace (upon request).\footnote{\url{https://huggingface.co/datasets/PRAIG/AMNLT}}

Each dataset has in principle the same ground truth setup: images of individual staffs, and their transcriptions, both of the melody and the text. Rather than following the older OMR pipelines based on object detection \cite{droettboom2003gamera,rebeloOMRStateOfTheArt2012,fujinaga2019single}, these are structured for end-to-end recognition, which has become a more dominant paradigm recently \cite{calvo-zaragozaEndtoEndNeuralOptical2018,rios-vilaEndtoendOpticalMusic2023,mayerPracticalEndToEnd2023,rios-vilaEndtoEndFullPageOptical2026}.

\begin{table}
    \begin{center}
        \begin{tabular}{l|l|l}
            \textbf{Dataset} & \# \textbf{samples} & \textbf{encoding} \\
            \hline
            GregoSynth & 126,579 & GABC \\
            Salzinnes & 2,940 & Pseudo-GABC \\
            Einsiedeln & 1,816 & Pseudo-GABC \\
            Solesmes & 854 & S-GABC \\
        \end{tabular}
    \end{center}
    \caption{Overview of available datasets along with their encoding format and number of samples. A ``sample'' in this case is one notated staff and its transcription.}
    \label{tab:datasets}
\end{table}

\subsection{GregoSynth}
\label{sec:gregosynth}

\begin{figure}[t]
    \centering
    \includegraphics[width=0.6\textwidth]{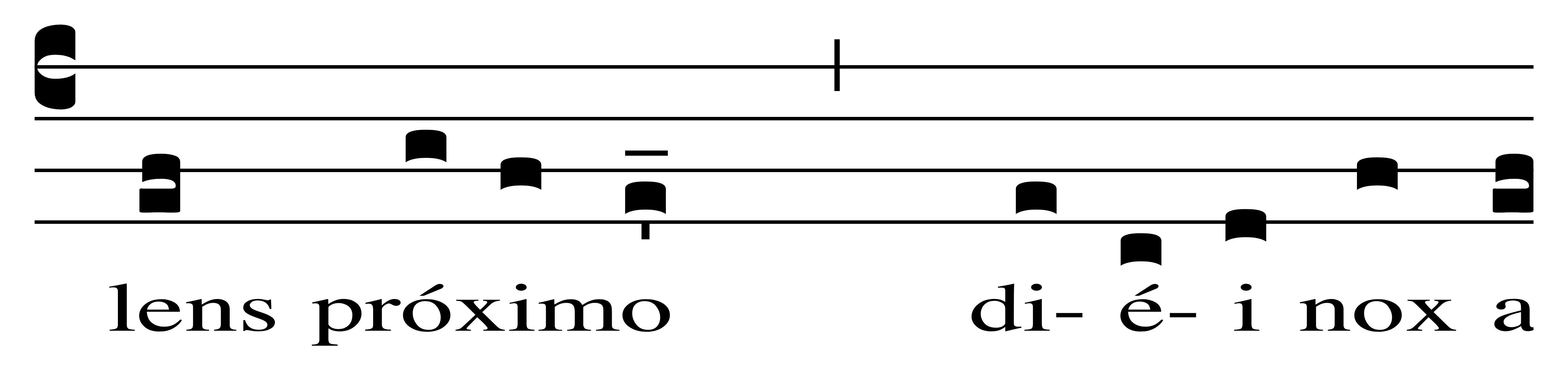}
    \begin{small}
    \begin{verbatim}
(<m>c<m>4)lens(<m>e<m>f) pró(<m>g)xi(<m>f)mo(<m>e<m>'<m>_) (<m>,)di(<m>e)
é(<m>c)i(<m>d) nox(<m>f) ad(<m>e<m>f)
    \end{verbatim}
    \end{small}
    \vspace{-2em}
    \caption{An example from GregoSynth dataset with its transcription encoded in music-aware GABC format.}
    \label{fig:gregosynth_example}
\end{figure}

The GregoSynth dataset, as the name suggests, is a dataset of synthetic Gregorian chants. It was generated from the Gregorian Chant Database\footnote{\url{https://gregobase.selapa.net/}} using GregorioTex\footnote{\url{https://gregorio-project.github.io/gregoriotex/}}. The transcriptions use GABC encoding with music aware tokenization (discussed in detail in Section~\ref{sec:encodings}). An example can be seen in figure \ref{fig:gregosynth_example}.

\subsection{Salzinnes}
\label{sec:salzinnes}

\begin{figure}[t]
    \centering
    \includegraphics[width=0.6\textwidth]{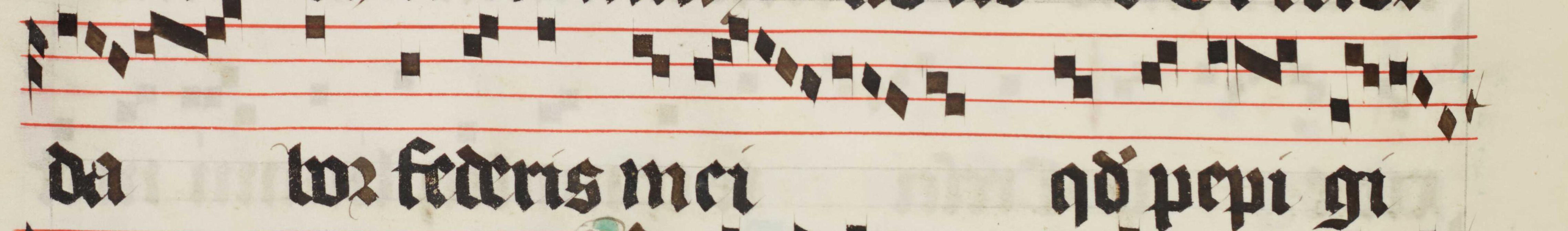}
    \begin{small}
    \begin{verbatim}
(C3) da(e3 d3-se c3-se d3 e3-l d3-l e3 f3-s) bor(e3-s) fe(c3) de(d3 e3) 
ris(e3-s) me(d3 c3) i(c3 d3 e3-s d3-se c3-se b2-se c3-s b2-se a2-se b2 a2) 
quod(c3 b2) pe(c3 d3-s) pi(d3-s d3-l c3-l d3-s) gi(a2 d3 c3 c3-s b2-se 
g2-se) z-a2
    \end{verbatim}
    \end{small}
    \vspace{-2em}
    \caption{Example from Salzinnes dataset with transcription encoded in Pseudo-GABC.}
    \label{fig:salzinnes_example}
\end{figure}

The Salzinnes dataset contains neumes from a 16th-century Cistercian antiphoner from the Abbey of Salzinnes, Namur, in the Diocese of Liège. It was originally transcribed in the Music Encoding Initiative (MEI) format; \cite{fuentes-martinezAlignedMusicNotation2026} converted the MEI format to Pseudo-GABC encoding (discussed in detail in section \ref{sec:encodings}). An example can be seen in figure \ref{fig:salzinnes_example}.

\subsection{Einsiedeln}
\label{sec:einsiedeln}

\begin{figure}[t]
    \centering
    \includegraphics[width=0.8\textwidth]{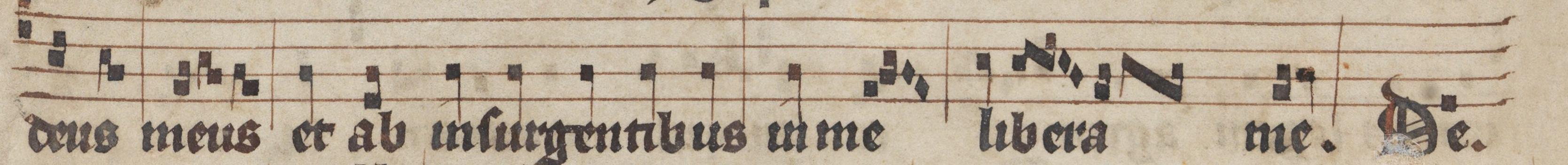}
    \begin{small}
    \begin{verbatim}
(C4) de(g2 a2) us(g2-n f2) | me(e2 f2 g2-n f2) us(f2-n e2) | et(f2-s) 
ab(d2 f2) in(f2-s) sur(f2-s) gen(f2-s) ti(f2-s) bus(f2-s) | in(f2-s) me(e2
f2 g2 f2-se e2-se) | li(g2-s) be(g2 a2-l g2-l a2 g2-se f2-se) ra(e2 f2 g2-l
e2-l f2) me(e2 f2 f2-s) | De(d2)
    \end{verbatim}
    \end{small}
    \vspace{-2em}
    \caption{Example from Einsiedeln dataset with transcription encoded in Pseudo-GABC.}
    \label{fig:einsiedeln_example}
\end{figure}

The Einsiedeln dataset contains neumes from a manuscript of the monastery of Einsiedeln, Switzerland. Its transcriptions are encoded the same as Salzinnes (section \ref{sec:salzinnes}), i.e. originally in MEI and then converted to Pseudo-GABC. An example can be seen in figure \ref{fig:einsiedeln_example}.

\subsection{Solesmes}
\label{sec:solesmes}

\begin{figure}[t]
    \centering
    \includegraphics[width=0.8\textwidth]{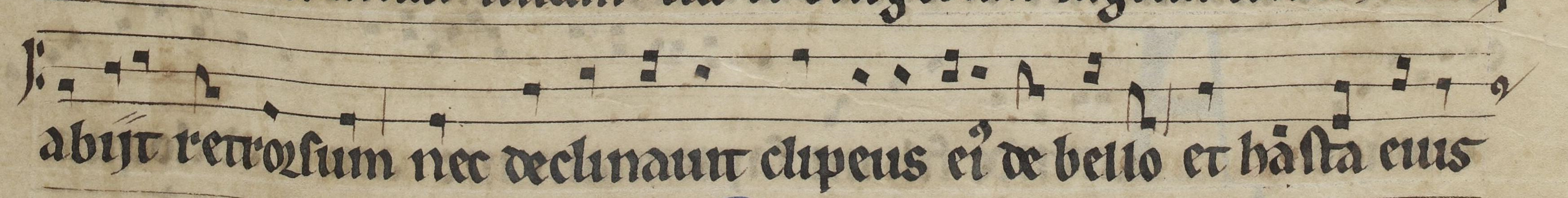}
    \begin{small}
    \begin{verbatim}
(<m>f<m>2) a(<m>e<m>v)bi(<m>f<m>v)it(<m>g<m>v) re(<m>f<m>e)tror(<m>D)sum
(<m>d<m>v) (<m>;) nec(<m>d<m>v) de(<m>f<m>v)cli(<m>g<m>v)na(<m>g<m>h)vit
(<m>g) cli(<m>h<m>v)pe(<m>G)us(<m>G) e(<m>g<m>h)i(<m>g) de(<m>g<m>f) bel
(<m>g<m>h)lo(<m>f<m>d) (<m>;) et(<m>f<m>v) ha(<m>c)sta(<m>d<m>f) e(<m>f
<m>g)ius(<m>f<m>v) (<m>f<m>+) ()
    \end{verbatim}
    \end{small}
    \vspace{-2em}
    \caption{An example from Solesmes dataset with its transcription encoded in S-GABC.}
    \label{fig:solesmes_example}
\end{figure}

The Solesmes dataset consists of Gregorian chants from the extensive collections of the Solesmes abbey that were central to the monks' early efforts at a critical edition of Gregorian chant in the early 20th century and since. The transcriptions are encoded in the S-GABC format introduced by Thomae et al. \cite{thomae2024preliminary}. An example can be seen in figure \ref{fig:solesmes_example}. This is the smallest and most diverse dataset, and therefore the most challenging of the four.

\subsection{Duplicates cleaning}

We found issues stemming from the fact that the underlying chants are present in multiple versions with only slight differences, which means that individual samples (staffs, which can also be relatively short) may contain the same (or extremely similar) ground truth string. This is a small issue in Salzinnes, Einsiedeln, and Solesmes, where the images corresponding to (near-)duplicate ground truth will still be different. In GregoSynth, however, the images are synthesised from the ground truth encoding, so they end up (near-)identical, to the pixel; this is a major issue especially when such duplicate sets end up distributed across both the training and test splits.

%I have performed a brief analysis on the datasets. I was looking for how similar are the samples to each other. We want to have as much variety in each dataset as possibly. Learning to transcribe the same sample with slight variations is not hard and the model will not be able to generalize. The analysis should also uncover contamination between splits which would lead to inaccurate results.

%In the first step, we focused on similarity between transcriptions, because of robustness and much lower computational complexity. 
We computed similarity between ground truth pairs as Levenshtein distance between the strings using the StringZilla library.\footnote{\url{https://github.com/ashvardanian/stringzilla}}
%developed by Ash Vardanian, accounting for the UTF-8 encoding.
%I have opted to use a distance calculation that correctly takes into account UTF-8 symbols\footnote{UTF-8 is a variable length encoding with up to 4 bytes per character}. 
%The transcriptions were preprocessed before computing the distance metric in order to detect only similarity between actual content and not irrelevant formatting instructions for the transcriptions (present in GABC). 
All datasets were first converted to the common encoding (see below), which removed most of the formatting symbols irrelevant to the content and with only slight implications for the visual appearance in the case of GregoSynth. We also removed all whitespace and music tags
as these do not change the content.

%I decided to clean the datasets of the duplicates. 
%There are might be several ways to define what to consider a duplicate and what not. We want to avoid having the same input and output pair multiple times in the dataset because it does not help the model to learn. It might be even lead to worse performance if some sample would be duplicated too many times. 

We then need to also consider image similarity. For example, the Salzinnes dataset has several samples with just a \textit{differentia} (lyrics ``\texttt{euoae}'') and the same melody which is normal. If it has exactly the same images, it is a true duplicate. However, if the images are visually different, we consider them to be two distinct samples, and they do not need to be discarded. We use the structural similarity index (SSIM) \cite{wang2004image}.%from scikit-image

The total definition of a duplicate is as follows: (1) the Levenshtein distance between samples is less than or equal to 3 and (2) the image similarity score (SSIM) is greater or equal 0.97. 

This pipeline was run on each dataset separately. The resulting dataset sizes can be found in table \ref{tab:dataset_cleaning}.

\begin{table}
    \begin{center}
        \begin{tabular}{|l|l|l|l|}
            \hline
            \multirow{2}{4em}{Dataset} & \multicolumn{3}{|c|}{\# samples} \\
            \cline{2-4}
            & train & validation & test \\
            \hline
            GregoSynth & 101,263 $\rightarrow$ \textbf{84,921} & 18,987 $\rightarrow$ \textbf{17,459} & 6,329 $\rightarrow$ \textbf{5,980} \\
            Salzinnes & 2,058 $\rightarrow$ \textbf{2,058} & 588 $\rightarrow$ \textbf{587} & 294 $\rightarrow$ \textbf{294} \\
            Einsiedeln & 1,271 $\rightarrow$ \textbf{1,271} & 363 $\rightarrow$ \textbf{363} & 182 $\rightarrow$ \textbf{182} \\
            Solesmes & 597 $\rightarrow$ \textbf{594} & 170 $\rightarrow$ \textbf{167} & 87 $\rightarrow$ \textbf{86} \\
            \hline
        \end{tabular}
    \end{center}
    \caption{Comparison of the number of samples in each split before and after the cleaning procedure. The final number of samples is highlighted in bold. Some samples were removed just because their transcription cannot be parsed using grammars implemented in this work.}
    \label{tab:dataset_cleaning}
\end{table}

%\textbf{TODO:Describe problems with the datasets.}
%\textbf{TODO:Describe deduplication process - probably using pseudo code or a numbered list}
%\textbf{TODO:Put a note somewhere about removed samples due to their encoding}

\section{Joint encoding}
\label{sec:encodings}

The four datasets are given in three different encodings: GABC, S-GABC, and a MEI-derived ``Pseudo-GABC''. So, despite being datasets pertaining to the same --- or, compatible enough --- type of notation in the images, the datasets cannot be combined, which is the opposite of helpful. 
To some extent, this even makes the prior results incomparable with each other: getting a lower error rate on Salzinnes than on Solesmes with the same model may not mean that Solesmes is harder to recognise, as the encoding of the Solesmes dataset may contain more symbols not entirely relevant to actually recovering text and melody (and aligning them, as framed in \cite{fuentes-martinezAlignedMusicNotation2026}) where errors are made. This is not to take issue with any particular encoding of chant: there are multiple good solutions to how to encode diastematic chant notation digitally. However, because the datasets contain images of the same tradition, with notation conforming to identical principles, there is no reason why the encodings couldn't be mapped onto each other.

This is the key step that we perform: we define a shared GABC-derived encoding to which all the datasets can then be re-encoded without losing information about melody, texts, their alignment, and grouping into neumes. We then re-encode the datasets in this shared encoding. The point of departure is the proposed Systematic GABC \cite{thomae2024preliminary}.

\subsection{GABC}

The GABC encoding was derived from ABC notation by the Gregorio project.\footnote{\url{https://gregorio-project.github.io/}; the full documentation for this encoding is part of Gregorio's github repository (\url{https://github.com/gregorio-project/gregorio})} %\todo{or is it just gregorio?}
The other encodings are derived from GABC; hence we describe it in more detail.

%The GABC notation also supports various features of Gregorian chants that do not appear in the early notations, e.g. \xxx{TODO}.
The encoding combines lyrics and notes in a single sequence. Since both share the same character set, they are distinguished by parentheses: musical notation is enclosed, while lyrics remain outside. The GregoSynth dataset has additionally differentiated the music notation by prepending \texttt{<m>} to every character, but only to differentiate music and lyric tokens in training models.

Notes are encoded as a single letter from \texttt{a} to \texttt{m} for the four line staff which is used in Gregorian chants. There is no separating character between notes. The letter describes the note position within the staff, not the pitch (in this, it is analogous to the Volpiano encoding used in the Cantus network of databases \cite{lacoste2012cantus,swanson2016volpiano}). 
%(figure \xxx{[...]}\todo{extract figure from gregorio documentation}). 
This puts a hard limit where we can place notes.

Each note can then have prefixes or suffixes that modify the visual appearance or the meaning of the note. Neither prefixes, nor suffixes have any separation character between them and the note.

The pitch of the notes are determined by a clef which is placed within the staff. There can be either a \texttt{c} clef or a \texttt{f} clef which symbolizes the position of a \texttt{c} or a \texttt{f} pitch. It can be modified to a flat clef which will symbolize \texttt{c} flat or \texttt{f} flat. The clef can be positioned only on a line which is indicated by a number from 1 to 4 in our case.

%\xxx{describe the GABC encoding}
%\xxx{mention Gregorio project}
%\xxx{describe the problems with the encoding - typos, nabc notation}

\subsection{Systematic GABC (S-GABC)}

The GABC encoding is intended for digital typesetting of modern Gregorian square notation used by the Solesmes editions of the 20th century. %(Specifically, GregorioTeX provides a conversion from GABC to TeX that can then in turn be rendered e.g. to a PDF.) 
The encoding thus supports various formatting options for lyrics and notes, 
%The negative side effect is that there can be a lot of characters in the encoding that only slightly changes the visual appearance and do not in fact encode anything relevant to the musical and textual content of the chant. This is
which are not used in other encodings that were designed with the opposite process (from image to structured encoding) in mind. %, and therefore they capture only the meaning of the notation and not its formatting (e.g. they do not encode the length of space between notes).
This use-case leads to the most important issue GABC has for the purposes of OMR: there are multiple ways to encode the same (or effectively the same) music.

To address this issue, Thomae et al. \cite{thomae2024preliminary} proposed Systematic GABC (S-GABC). This canonical encoding formalizes the GABC definition by context-free grammar which avoids any ambiguities of the original encoding. However, the proposed grammar is not complete, and some rules are not fully defined. Additionally, the ground truth of the Solesmes dataset, which is encoded in S-GABC, does not fully comply with the specification in \cite{thomae2024preliminary}. 
%I had to make some educated guesses in order to write a grammar that parses most of the dataset's transcriptions.

Systematic GABC has the same encoding style as GABC, differing only in a few things. Hollow notes are not supported; instead, the letter \texttt{r} is used to encode uncertain reading of neumes. Completely illegible sections are marked with empty parentheses. Irrelevant options for formatting lyrics are not supported. 
%
%These are just the major differences; for detailed comparison see table ... \textbf{TODO:xxx[...] create table or maybe attachment would be better since it is quite long}

Like GregoSynth, the Solesmes dataset also has the music tag \texttt{<m>} for every music character.

%\xxx{describe the S-GABC encoding}
%\xxx{describe the problems with the encoding - original grammar is not documented sufficiently, some typos}

\subsection{Pseudo-GABC}

Finally, there is the Pseudo-GABC encoding. This was introduced in \cite{fuentes-martinezAlignedMusicNotation2026} for data converted from the XML-based Music Encoding Initiative (MEI) neumes format to a linear structure similar to the GABC encoding. However, Pseudo-GABC uses absolute pitches as note symbols instead of ``agnostic'' position on the staff lines. 

%Besides not being compatible with the GABC and S-GABC encodings, this has the issue (well-known in OMR evaluation \cite{byrdStandardTestbedOptical2015,calvo-zaragozaUnderstandingOpticalMusic2021,torrasUnifiedRepresentationFramework2024,martinez-sevillaSheetMusicBenchmark2025b}) that predicting the wrong clef results in all the pitches being wrong as well; from another perspective, it makes the relationship between a part of the input image with one note and the correct output symbol depend also on another part of the input image.
%subset of GABC has the same downside if we care only about the absolute pitches of the notes. 
%However, the Levenshtein distance is much smaller in the GABC style encoding for this error. So, if we were to correct the mistakes, the GABC style encoding has the advantage in less changed characters.

Pseudo-GABC is, unfortunately, not documented. However, it was possible to reverse-engineeer the meanings of the encoding symbols from the Salzinnes and Einsiedeln datasets, in combination with the documentation of MEI Neumes module.

Pseudo-GABC, like GABC, encloses all melody encoding in parentheses, and everything outside is lyrics. Notes are encoded as absolute pitches, with optional suffixes indicating neume shapes (e.g., the presence of a stem). The pitch is always a pitch in English format (\texttt{c}-\texttt{b}) and a number that represents the octave.
%, e.g. \texttt{f2} = pitch \texttt{f} in the second octave.\footnote{This nomenclature is according to the German musicological tradition, where the central C (MIDI code 60) is called ``c1'', the octave above that is ``c2'', ``c3'', etc., and below is ``c'', ``C'', and ``C1''.} 
Suffix is separated from the pitch or other suffix by dash (\texttt{-}). 
%The is indicated by one or two letters.
The clef is encoded as in GABC: a letter indicating a clef type, and a number denoting the line on which it is placed. However, the clef type is an uppercase letter instead of lowercase one.
Accidentals are encoded as a symbol \texttt{f} for flat or \texttt{n} for neutral before a pitch, with a single space between them.

%\xxx{describe the PSEUDO GABC encoding in more details}
%\xxx{describe the problems with the encoding - typos}

\subsection{Defining a joint encoding and the conversion process}

Based on a thorough comparison of the features in the GABC, S-GABC, and Psuedo-GABC encodings used in the four available datasets, we developed a common encoding.
The main criterium was retaining all essential information about the melody and text as possible while keeping the encoding consistent across datasets. For example, neume spacing was removed, because Pseudo-GABC does not support it and it cannot be inferred unambiguously. A special marking for the \textit{porrectus} neume with an oblique component was kept, however, because while GABC did not support it, it was possible to write an algorithm which added the oblique marking retroactively. 
%
%Technical details are available in the project's GitHub.\footnote{\url{https://github.com/danielkurek/GabcParser}}
The full list of supported symbols across the three input encodings and the common encoding is provided.\footnote{\url{https://github.com/danielkurek/GabcParser/blob/main/comparison_table.md}}

The conversion to the common encoding\footnote{\url{https://github.com/danielkurek/GabcParser/blob/main/gabcparser/utils/common_encoding.py}} is implemented as a set of Lark grammars for each of the GABC,\footnote{GregorioTex v. 6.1.0 was used to create the context-free grammar for parsing GABC.} S-GABC, and Pseudo-GABC encodings, and an output CFG into the shared format. The transcriptions are parsed using corresponding grammar into a parse tree. The parse tree is then transformed to a tree that represents the common encoding\footnote{The transformed parse tree structure might differ slightly from the one that is generated when parsing with the common grammar.}.
Due to malformed strings in the original datasets, not all samples were successfully re-encoded, though not more than 5 \% of each dataset. The full report of mistakes is made available.\footnote{\url{https://github.com/danielkurek/GabcParser/blob/main/README.md}}

%\xxx{compare the features of individual datasets}
%\xxx{how was the common encoding chosen}
%\xxx{}

\begin{table}[ht!]
    \centering
    \bgroup
    \def\arraystrech{2.0}
    \setlength\tabcolsep{4pt}
    \begin{tabular}{lrrrrr}
        \toprule
         \textbf{Model} & \textbf{MER} & \textbf{CER} & \textbf{SylER} & \textbf{ALER} & \textbf{AMLER} \\
         \midrule
         \textbf{GregoSynth} & & & & & \\
         %\arrayrulecolor{black!20}\midrule         
         Prev.best (D.\&C./SMT) & 2.26 & 4.54 & 8.66 & \textbf{0.09} & 2.93 \\
         DAN-baseline  & 9.05 & 11.74 & 25.00 & 0.16 & 9.52 \\
         %\arrayrulecolor{black!20}\midrule
         DAN-shared  & 5.42 & 5.8 & 13.79 & 0.19 & 6.22 \\
         %DAN-FT  & -- & -- & -- & -- & -- & -- \\
         DAN(8)-shared  & \textbf{1.45} & \textbf{2.56} & \textbf{6.91} & \textbf{0.09} & \textbf{2.23} \\
         \arrayrulecolor{black}\midrule
         \textbf{Salzinnes} & & & & & \\
         %\arrayrulecolor{black!20}\midrule         
         Prev.best (SMT/D.\&C.)  & 13.73 & \textbf{\textit{2.84}} & 7.80 & \textbf{0.14} & 7.32 \\ 
         DAN-baseline  & 29.46 & 12.34 & 21.08 & 0.25 & 18.73 \\         
         %\arrayrulecolor{black!20}\midrule
         DAN-shared  & 11.63 & 9.76 & 18.54 & 0.22 & 11.52 \\
         DAN-FT  & 7.34 & 3.68 & 8.26 & 0.18 & 6.33 \\
         DAN(8)-shared  & \textbf{\textit{6.61}} & \textbf{\textit{2.61}} & 5.75 & \textbf{\textit{0.15}} & \textbf{\textit{5.39}} \\
         DAN(8)-FT  & \textbf{6.55} & \textbf{2.56} & \textbf{5.15} & 0.16 & \textbf{5.18} \\
         \arrayrulecolor{black}\midrule     
         \textbf{Einsiedeln} & & & & & \\ 
         %\arrayrulecolor{black!20}\midrule
         Prev.best (=DAN-baseline)  & 7.23 & 3.44 & 7.19 & \textbf{0.17} & 5.14 \\
         %DAN-baseline  & \ditto & \ditto & \ditto &  \ditto & \ditto & -- \\ 
         %\arrayrulecolor{black!20}\midrule
         DAN-shared  & 14.64 & 13.99 & 24.72 & 0.35 & 15.14 \\
         DAN-FT  & 3.8 & 4.39 & 8.59 & 0.22 & 4.26 \\
         DAN(8)-shared  & 3.39 & 3.63 & 7.09 & 0.22 & 3.8 \\
         DAN(8)-FT  & \textbf{2.33} & \textbf{2.75} & \textbf{5.81} & 0.25 & \textbf{2.73} \\
         \arrayrulecolor{black}\midrule
         \textbf{Solesmes} & & & & & \\  
         %\arrayrulecolor{black!20}\midrule         
         Prev.best (D.\&C./CRNN)  & 17.72 & 8.86 & 20.84 & 0.53 & 20.98 \\
         DAN-baseline  & 21.87	& 17.94	& 30.19 & 0.39 & 23.17 \\     
         %\arrayrulecolor{black!20}\midrule         
         DAN-shared  & 30.82 & 30.8 & 49.21 & 0.42 & 31.08 \\
         DAN-FT  & 18.63 & 14.08 & 25.56 & 0.33 & 18.63 \\
         DAN(8)-shared  & 12.35 & 9.9 & 18.6 & \textbf{0.31} & 13.07 \\
         DAN(8)-FT  & \textbf{10.25} & \textbf{7.79} & \textbf{16.07} & \textbf{0.31} & \textbf{10.81} \\
         \arrayrulecolor{black}\bottomrule
    \end{tabular}
    \egroup
    \vspace{4pt}
    \caption{Main experimental results. Baseline result on each dataset both for vanilla DAN for direct comparison, and for the previously best-performing model in each metric separately.%, to show the combined influence of individual model steps. 
    %BWER for baselines is computed by inverting the AMLER to ALER formula, as AMLER x (1 - ALER). 
    Models we trained are DAN-shared (combined data, balanced, no fine-tuning), DAN-FT (fine-tuning DAN-shared on individual datasets; not performed on GregoSynth), DAN(8)-shared (combined data, 8-sample gradient accumulation), and DAN(8)-FT (fine-tuning DAN(8)-shared on individual datasets, 8-sample gr. acc.; not performed on GregoSynth). Best-performing model in each metric on each dataset in bold; close second-bests (within 0.5, or 0.01 in ALER) italicised bold.}
    \label{tab:results}
\end{table}

\section{Recognition experiments and Results}

To measure the potential of pooling datasets, we train a shared model. Out of the architectures used in \cite{fuentes-martinezAlignedMusicNotation2026}, we choose the DAN model \cite{Coquenet2023DAN}: it performed well, and in preliminary experiments, it was faster to train than the Sheet Music Transformer. For the combined training, we randomly downsample GregoSynth in each epoch to the same number of training examples as second the largest dataset (that is, Salzinnes), so that the synthetic data don't overwhelm the training process while taking advantage of its more varied musical content.

To train the shared model, we copy the setup of \cite{fuentes-martinezAlignedMusicNotation2026}. To see the the potential of the shared model as a foundational model, we then run fine-tuning experiments, where we use the same training procedure on each of the real datasets (Salzinnes, Einsiedeln, Solesmes) initialised with the weights of the shared model with learning rate of $1e^{-5}$ instead of the original $1e^{-4}$.

The minibatches in \cite{fuentes-martinezAlignedMusicNotation2026} are of size 1 due to uneven scales of input images. Noting this as a potential weakness in the training setup, we also train the shared model with 8-fold gradient accumulation instead.

We retain the train/validation/test splits for all four datasets as in \cite{fuentes-martinezAlignedMusicNotation2026}. The encodings changed between the baseline models and models trained in this work, so the comparison between baseline and our models is not quite direct; however, the vast majority of symbols participating in the evaluation metric are retained, so at least an approximate comparison is still meaningful. %; a more substantial difference might be in the Salzinnes and Einsiedeln datasets.

%\section{Results}
We report the same metrics as \cite{fuentes-martinezAlignedMusicNotation2026}: symbol error rates on the subset of symbols that comprise the melody (MER), the sung text (CER), entire combined syllables (SylER), the percentage of errors caused by melody-text alignment issues (ALER), and the combined rate of all errors in melody, text, and alignment together (AMLER). %and finally the combined error rate for melody and text but disregarding alignment issues (BWER, for Bag-of-Words Error Rate). ALER is computed as the ratio of BWER to AMLER. 
For applications in an automated manuscript cataloguing and retrieval (searh), CER and MER are most important; for use in workflows that produce inputs for human readers, the combined AMLER is most indicative of overall applicability.
All results are reported in Table~\ref{tab:results}.

\section{Discussion and Conclusions}

What did harmonising the encodings and cleaning the datasets achieve? 

The DAN-shared model does not match the best baseline across all datasets, but at least on GregoSynth and Salzinnes, it outperforms DAN-baseline. This is not entirely unexpected: the shared model has a somewhat harder job generalising across all four dataset (though it saw training samples from them all). 

However, the real impact of harmonising the encodings is shown to be enabling the shared pretraining + fine-tuning pipeline. The fine-tuned DAN models match or exceed the performance of the best baseline results (except for Solsemes CER). This confirms the value of merging datasets when the underlying notation is mutually compatible, as already observed for white mensural notation \cite{martinez-sevillaTowardsUniversalOptical2024}.

But then the shared model with 8-fold gradient accumulation outperforms \textit{everything else}. While a complete comparison would involve training the DAN baseline with the same gradient accumulation policy rather than a batch size of 1,\footnote{See: \url{https://github.com/efm18/AMNLT/tree/main/configs/smt_dan_config/Solesmes}} the shared model might be able to take advantage of gradients combined across samples from different datasets (GregoSynth was undersampled in every epoch), thus starting to take advantage of the greater diversity of notations in the combined datasets. Fine-tuning this model then improves the results further, though percentage-wise the impact is not as large as before: the improved training allows already the shared model to learn more from the data it had. %The achieved CER and MER on Solesmes data may be good enough at least for automated cataloguing of chant manuscripts. %Experiments with fine-tuning this model are ongoing; the most visually diverse (and smallest) dataset, Solesmes, still has MER over 12.0, likely making the shared model not quite good enough for practical application.

The common encoding is very much based on the S-GABC proposal, which was a first step in this direction of a shared diastematic chant notation encoding; however, there are some differences from S-GABC as well: all lyric formatting was removed, since it was not used in datesets, uncertain readings were removed, and hollow notes were reintroduced using the same syntax as GABC. Horizontal episema position was reduced to only 0 and 1. We intentionally do not give a name for our encoding, because we believe it could simply be used together with the original S-GABC proposal as the next step towards a full implementation of what S-GABC could be, and we would welcome a discussion on the future.

In any case, the effort put into the harmonising encodings and cleaning the datasets has resulted in a new state of the art for end-to-end OMR that is packaged into a more universal model that can provide these results across the diverse visual domains of Gregorian notations. 

%\section*{Acknowledgements}

\begin{credits}
\subsubsection{\ackname}%
%This work has been supported by the Ministry of Culture of the Czech Republic (project OmniOMR of the NAKI III programme, no. DH23P03OVV008). The computing infrastructure was provided by the LINDAT/CLARIAH-CZ Research Infrastructure,\footnote{\url{https://lindat.cz}} supported by the Ministry of Education, Youth and Sports of the Czech Republic (project no. LM2023062).
This work has been supported by the Ministry of Culture of the Czech Republic (project OmniOMR of the NAKI III programme, no. DH23P03OVV008), the project ``Human-centred AI for a Sustainable and Adaptive Society'' (reg. no.: CZ.02.01.01/00/23\_025/0008691), co-funded by the European Union, and the Digital Analysis of Chant Transmission (DACT) project, funded by a Partnership Grant from the Social Sciences and Humanities Research Council of Canada (895-2023-1002). The computing infrastructure was provided by the LINDAT/CLARIAH-CZ Research Infrastructure,\footnote{\url{https://lindat.cz}} supported by the Ministry of Education, Youth and Sports of the Czech Republic (project no. LM2023062).

%\subsubsection{\discintname}
%It is now necessary to declare any competing interests or to specifically state that the authors have no competing interests. Please place the statement with a bold run-in heading in small font size beneath the (optional) acknowledgments, for example: 
%The authors have no competing interests to declare that are relevant to the content of this article. 
%Or: Author A has received research grants from Company W. Author B has received a speaker honorarium from Company X and owns stock in Company Y. Author C is a member of committee Z.
\end{credits}

% \section{Author Contributions}

% (Redacted for peer review.)

% \noindent
% \textbf{Vojtěch Dvořák:} Formal analysis, Investigation, Software, Validation, Visualization, Writing – review \& editing.

% \noindent
% \textbf{Filip B\v{i}m:} Data curation, Investigation, Methodology, Project administration, Resources, Validation, Writing – review \& editing.

% \noindent
% \textbf{Ji\v{r}\'{i} Mayer:} Conceptualization, Data curation, Formal analysis, Investigation, Methodology, Project administration, Software, Validation, Visualization, Writing – original draft, Writing – review \& editing.

% \noindent
% \textbf{Martina Dvo\v{r}\'{a}kov\'{a}:} Data curation, Investigation, Methodology, Project administration, Resources, Validation, Writing – review \& editing.

% \noindent
% \textbf{Markéta Herzánová Vlková:} Data curation, Investigation, Methodology, Project administration, Resources, Validation, Writing – review \& editing.

% \noindent
% \textbf{Petr Žabička:} Data curation, Funding acquisition, Project administration, Resources, Supervision, Writing – review \& editing.

% \noindent
% \textbf{Jan Hajič jr.:} Conceptualization, Funding acquisition, Methodology, Project administration, Resources, Supervision, Visualization, Writing – original draft, Writing – review \& editing.

% Conceptualization
% Data curation
% Formal analysis
% Funding acquisition
% Investigation
% Methodology
% Project administration
% Resources
% Software
% Supervision
% Validation
% Visualization
% Writing – original draft
% Writing – review & editing

\bibliography{bibliography}
\bibliographystyle{ieeetr}

\end{document}